\documentclass[runningheads]{llncs}
\usepackage[10pt]{extsizes}
\usepackage{graphicx}
\usepackage{algorithm}
\usepackage{algorithmic}
\usepackage{amsmath}
\usepackage{booktabs}
\usepackage{dsfont}
\usepackage{blindtext} 
\usepackage{array,booktabs}
\newcolumntype{L}{@{}>{\kern\tabcolsep}l<{\kern\tabcolsep}}
\usepackage{colortbl}
\usepackage{xcolor}
\usepackage{hyperref}
\usepackage{cleveref}
\definecolor{maroon}{rgb}{0.5, 0, 0}
\definecolor{forestgreen}{rgb}{0.28, 0.49, 0.17}

\definecolor{dazzledblue}{RGB}{52, 89, 149}

\usepackage{soul}

\begin{document}
\title{Protoformer: Embedding Prototypes for Transformers}

\author{Ashkan Farhangi\thanks{Corresponding author.} \and
Ning Sui \and
Nan Hua \and 
Haiyan Bai
\and  Arthur Huang
\and Zhishan Guo
} 
\authorrunning{A. Farhangi et al.}

\institute{University of Central Florida, Orlando FL, USA \\
\email{ashkan.farhangi@ucf.edu, zhishan.guo@ucf.edu}\\}
\maketitle              
\begin{abstract}
Transformers have been widely applied in text classification.
Unfortunately, real-world data contain anomalies and noisy labels that cause challenges for state-of-art Transformers. 
This paper proposes Protoformer, a novel self-learning framework for Transformers that can leverage problematic samples for text classification. Protoformer features a selection mechanism for embedding samples that 
allows us to efficiently extract and utilize anomalies prototypes and difficult class prototypes. 
We demonstrated such capabilities on datasets with diverse textual structures (e.g., Twitter, IMDB, ArXiv). We also applied the framework to several models. The results indicate that Protoformer can improve current Transformers in various empirical settings.
\keywords{Text Classification \and  Twitter Analysis \and Class Prototype}
\end{abstract} 

\section{Introduction}
\vspace{-2mm}
For real-world textual datasets, anomalies are known as samples that depart from the standard samples. Such anomalies tend to have scattered textual distributions, which can cause performance drops for state-of-art Transformer models~\cite{vaswani2017attentionTransformersOG}.
Moreover, models that rely on supervised learning can suffer from incorrect convergence when provided with noisy labeled data gathered from Internet~\cite{wei2020combating}.
Hence, there is a need to automatically detect the anomalies and adjust noisy labels to make the model more robust to complex noisy datasets.

As human annotations can be highly time-and-cost inefficient, it is more common that noisy labels are gathered from the Internet. For instance, Twitter has been increasingly adopted to understand human behavior~\cite{fiok2021Maham}. However, such data tend to complex and often contain noisy labels. This can make the standard supervised learning objective lead to incorrect convergence~\cite{garg2020noisy}.

One of the applications of this study is to classify college students' academic major choices based on their historical Tweets. When students follow a certain college's official account, it might indicate that the student belongs to that major. However, there are uncertainties about the correctness of the labels.
Therefore, the supervised model's results can become untrustworthy.

\begin{figure}[!t]
 \centering
 \includegraphics[trim={1.7cm 2.6cm 4.2cm 2.2cm},clip,width =.75\linewidth]{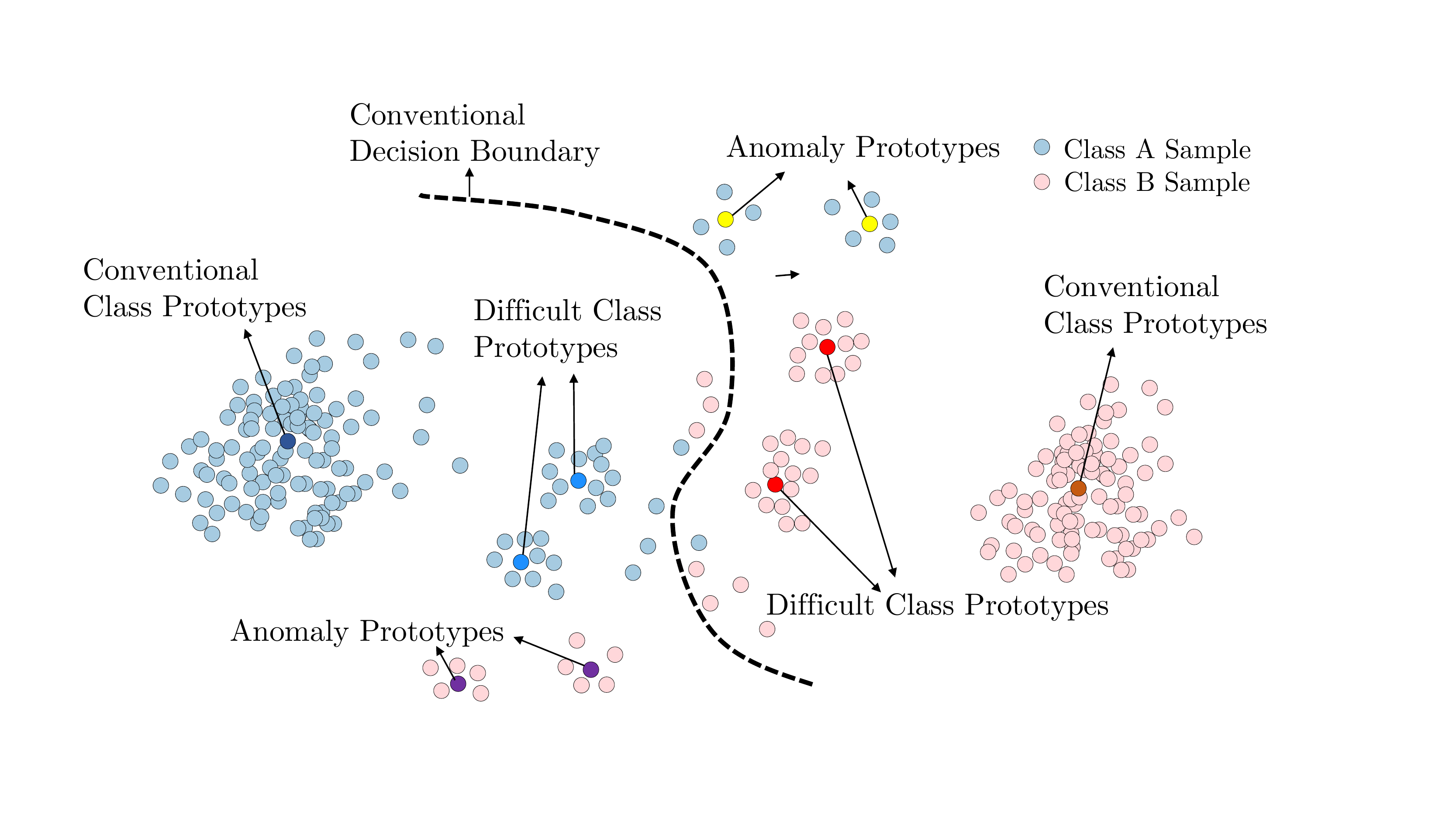}
 \caption{Distribution of embeddings for real world data samples is often scatterd. Although conventional class prototypes are easier to select, difficult class prototypes and anomaly prototypes require a more careful approach in selection and play a critical role in improving the decision boundary.}
 \label{fig:Intro}
  \vspace{-5mm}

\end{figure}

There are some prior works on prototype embeddings. CleanNet~\cite{lee2018cleannet} proposes providing extra supervision for the training. Subsequently, SMP~\cite{han2019noisy} proposes using multiple prototypes to capture embeddings with high density without extra human supervision. However, both approaches do not provide a solution for troublesome embeddings that are scattered and are often minorities, as shown in Figure~\ref{fig:Intro}.
To alleviate this issue, we select prototypes through their contextual embeddings in a way to not only cover the difficult-to-classify samples but also represent minority samples of the dataset (i.e., anomalies).

We propose Protoformer framework that selects and leverages multiple embedding prototypes to enable Transformer's specialization ability to classify noisy labeled data populated with anomalies.
Specifically, we improve the generalization ability of Transformers for problematic samples of a class through \textit{difficult class prototypes} and their specialization ability for minority samples of a class through \textit{anomaly prototypes}. We show that the representations of both prototypes are necessary to improve the model's performance. Protoformer leverages these prototypes in a self-learning procedure to further improve the robustness of textual classification. To our best knowledge, this is the first study that extracts and leverages anomaly prototypes for Transformers.

In summary, the contributions are threefold:

\noindent $\bullet$ We propose a novel framework that learns to leverage harder to classify and anomaly samples. This acts as a solution for classifying datasets with complex samples crawled from the Internet. 
 
\noindent $\bullet$ The framework contains a label adjustment procedure and thus is robust to noise. This makes the framework suitable for noisy Internet data and can be used to promote a more robust Transformer model. Leveraging the similarity in the embedding space and a ranking metric, we can identify questionable labels and provide a certain level of adjustment. This mitigates the potential negative impact on the training.

\noindent $\bullet$ We evaluate the framework based on multiple datasets with both clean and noisy labels. Results show that our model improves the testing accuracy from 95.7\% to 96.8\% on the IMDB movie review dataset. For a self-gathered Twitter dataset with noisier labels, the classification accuracy improved with a greater margin  (from 56.7\% to 81.3\%).
\medskip

\section{Problem Formulation}
\label{section-preliminary}
Given a sample text as $\mathbf{x}_{i}$, $\mathbf{X}=\left\{\mathbf{x}_{1}, \mathbf{x}_{2}, \cdots, \mathbf{x}_{N}\right\}$ represents all the $N$ samples of the dataset, while $\hat{Y}=\left\{\hat{y}_{1}, \hat{y}_{2}, \cdots, \hat{y}_{N}\right\}$ indicates the corresponding noisy labels from the Internet.
The noisy label $\hat{y}_{i} \in \{0,1\}^{\Bar{c}}$ is a binary vector format with only one non-zero element, indicating the class label of
$\mathbf{x}_{i}$, where $\Bar{c}$ is the total number of classes.
A Transformer model $\mathcal{F}_W$ can be used as a classification model to produce an estimated label $\mathcal{F}_W(\mathbf{x}_{i}) \in [0,1]^{\Bar{c}}$, where $W$ represents the parameters.
The optimization strategy is based on the cross-entropy loss function:
\begin{equation}
\mathcal{L}(\mathcal{F}_{W}(\mathbf{x}_i), \Tilde{y}_i)=
-\sum_{j=1}^{\Bar{c}} \Tilde{y}_{i,j} \log \left(\mathcal{F}_{W}(\mathbf{x}_i)_j\right),
\end{equation}

In addition, labels from the internet are often noisy. Hence, as detailed in Section \ref{sec:noisy}, the labels can be adjusted according to the similarities of the class prototypes, resulting in adjusted labels $\Tilde{y}_i\in [0,1]^{\Bar{c}}$---it is a probability distribution, and thus $\sum_{j=1}^{\Bar{c}} \Tilde{y}_{i,j} = 1$. Even when we have sufficient confidence in the original labels, we can use it as a complementary supervision.

Specifically, for each batch with $m$ samples, we would pursue the following optimization problem:
\begin{equation}
W^{*}=\operatorname{argmin}_{W} \frac{1}{m}\sum_{i=1}^m \mathcal{L}\left(\mathcal{F}_{W}(\mathbf{x}_i), \Tilde{y}_i \right)
\end{equation}

\section{Design of Protoformer}
\vspace{-2mm}
\label{section-methodology}
This section provides the details of Protoformer. Specifically, we describe a procedure for extracting the difficult class prototypes (Section \ref{subsec:cp}). Subsequently, we describe a procedure for extracting anomaly prototypes (Section \ref{subsec:ap}). Both types of prototypes are then used in a multi-objective self-learning training process that optimizes the network parameters for robust text classification (Section \ref{subsec:self-learning}).
In order to handle noisy labeled data, we adjust the noisy labels through a label adjustment procedure that uses the prototype similarities (Section \ref{sec:noisy}). 

\subsection{Difficult Class Prototypes}
\label{subsec:cp}
\begin{figure}[!t]%
 \centering
 {{\includegraphics[trim={6.5cm 2.5cm 9cm 2cm},clip,width =.5\linewidth]{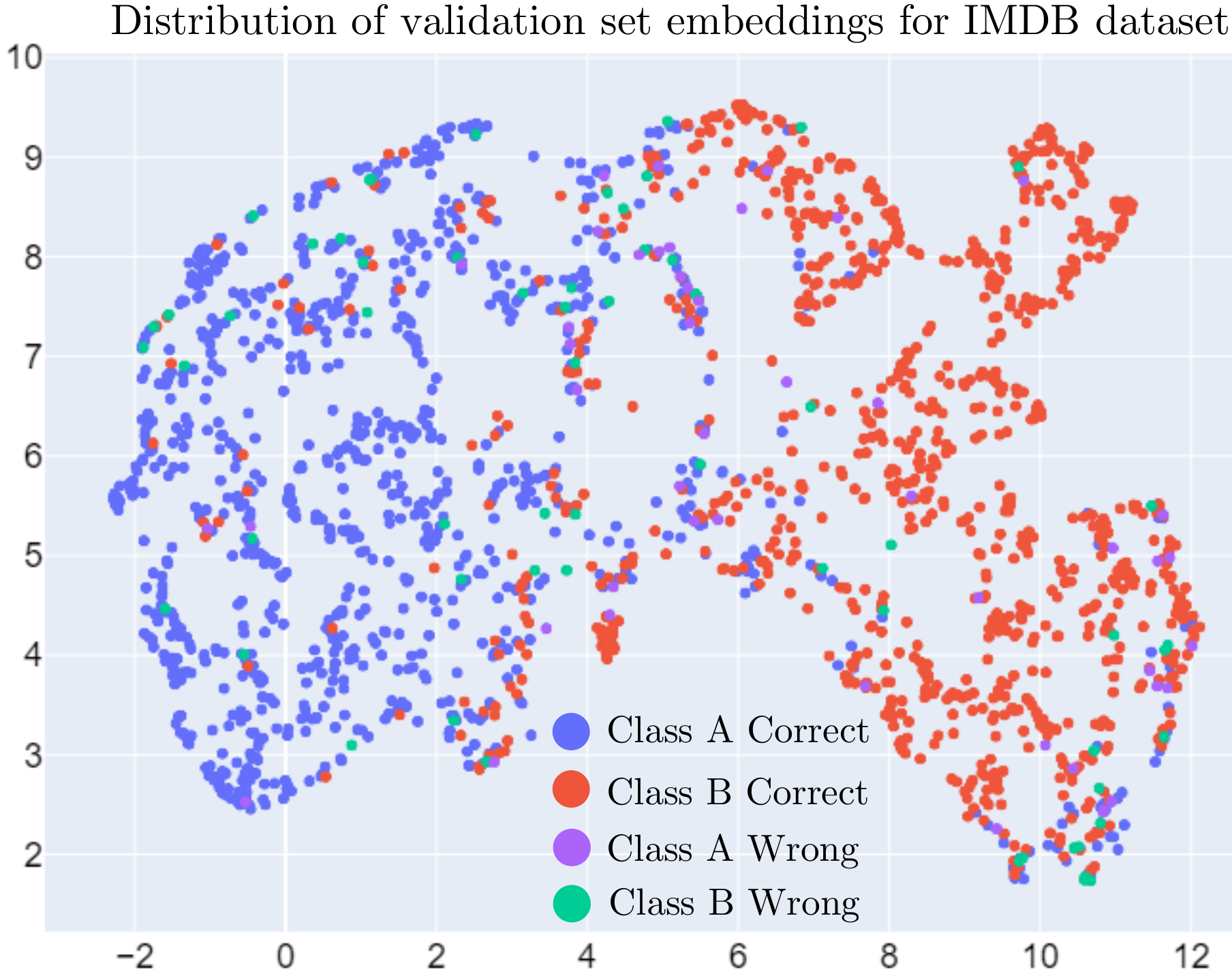}}}%
 \qquad
 {{\includegraphics[width=.445\linewidth]{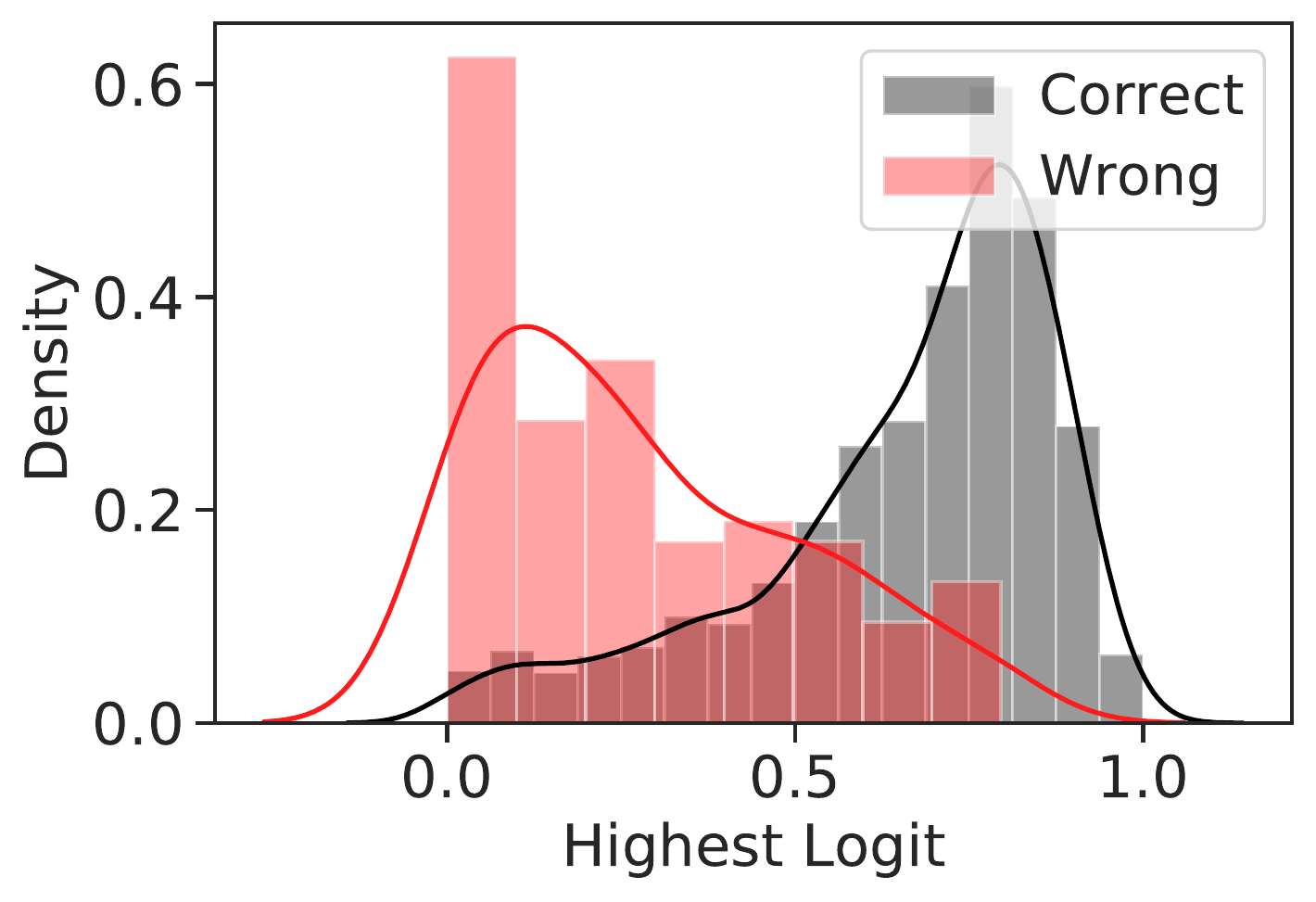} }}%
 \caption{\textbf{Left:} distribution of the embedding for IMDB dataset. Presence of anomalies aand problematic samples cause misclassification. \textbf{Right:} Distribution of the highest output logits (scaled 0-1) for the same model. 
 The higher values of the largest logits can represent the confidence of the network's classification.}
 \label{fig:accuracy_improve}
 \vspace{-5mm}

\end{figure}

Difficult class prototypes act as the representatives for the problematic samples of the dataset. For example, Figure~\ref{fig:accuracy_improve} showcases the fine-tuned embeddings of a benchmark dataset gathered from the Internet (i.e., IMDB). Although the majority of samples of each class are located closely together, there are anomaly samples that are scattered and often far from the majority. Unfortunately, these harder-to-classify samples are not the target focus of the state-of-art models in text classification. 
Moreover, traditional clustering methods (e.g., K-means) are not designed to capture or cluster such samples that are scattered and distributed throughout the embedding space. 

Intuitively, these problematic samples can cause the greatest error.
For instance, Figure~\ref{fig:accuracy_improve} also shows the classification error of the fine-tuned BERT~\cite{devlin2018bert} model where the majority of the classification error stems from harder-to-classify samples (over 51\%).
Such error arises when the highest classification logit values are still low and in between classes, which indicates the indecisiveness of the Transformer. 
Following \cite{han2019noisy}, we define the similarity of the extracted embeddings through {\bf pairwise similarity score} (i.e., cosine distance) of any two inputs $\mathbf{x}_i$ and $\mathbf{x}_j$ as:
\begin{equation}
s_{i j} = \frac{ \mathbf{e}\left(\mathbf{x}_{i}\right)^{T}\cdot \mathbf{e}\left(\mathbf{x}_{j}\right)}{\left\|\mathbf{e}\left(\mathbf{x}_{i}\right)\right\|_2\left\|\mathbf{e}\left(\mathbf{x}_{j}\right)\right\|_2},
\end{equation}
where $\mathbf{e}(\mathbf{x})$ is the embedding vector of sample $\mathbf{x}$, extracted from the first layer of the Transformer\footnote{For large-scale datasets, one can randomly choose a limited number
(e.g., $q$) 
of samples per class to develop a triangular similarity matrix $S^{q\times q}$ which can enhance the computational efficiency.}.

To determine the closeness of embeddings, we also define the {\bf proximity} metric $p$ for each embedding as:
\begin{equation}
p_{i}=\sum_{j=1}^{m} sign\left(s_{i j}-s_{c}\right),
\end{equation}
where $sign(x)$ is a sign function\footnote{$sign(x) = 1$ for $x > 0$, $sign(x) = 0$ for $x = 0$, and $sign(x) = -1$ otherwise.} 
and $s_c$ is an arbitrary value from the similarity matrix (default as 20-percentile).
Intuitively, a higher proximity indicates that the textual embeddings have more similar embeddings around them and are `closer' to every other sample in the embedding space.

Follwing \cite{pleiss2020noisyAUM}, problematic samples cause low confidence in output logits of the model. Hence, we define the \textbf{confidence} metric $c$ as:
\begin{equation}
 c_{i} = |\overbrace{\operatorname{max}_{\hat{c}^{1}}{
 \mathcal{F}_{W}(\mathbf{x}_i)_{\hat{c}^{1}}}}^{\text {largest logit}}-\overbrace{\operatorname{max}_{\hat{c}^{2}}
 \mathcal{F}_{W}
 (\mathbf{x}_i)_{\hat{c}^{2}}}
 ^{\text{second largest logit}}|
\end{equation}
where logits are scaled (0-1 range) and are taken from the output before the softmax layer after a preliminary training stage. 
Intuitively, when the confidence is low (near zero), the model indecisivess is the highest.
\begin{figure}[!t]%
 \centering\offinterlineskip
{{\includegraphics[trim={8cm 4cm 13.5cm 3cm},clip,width =.5\linewidth]{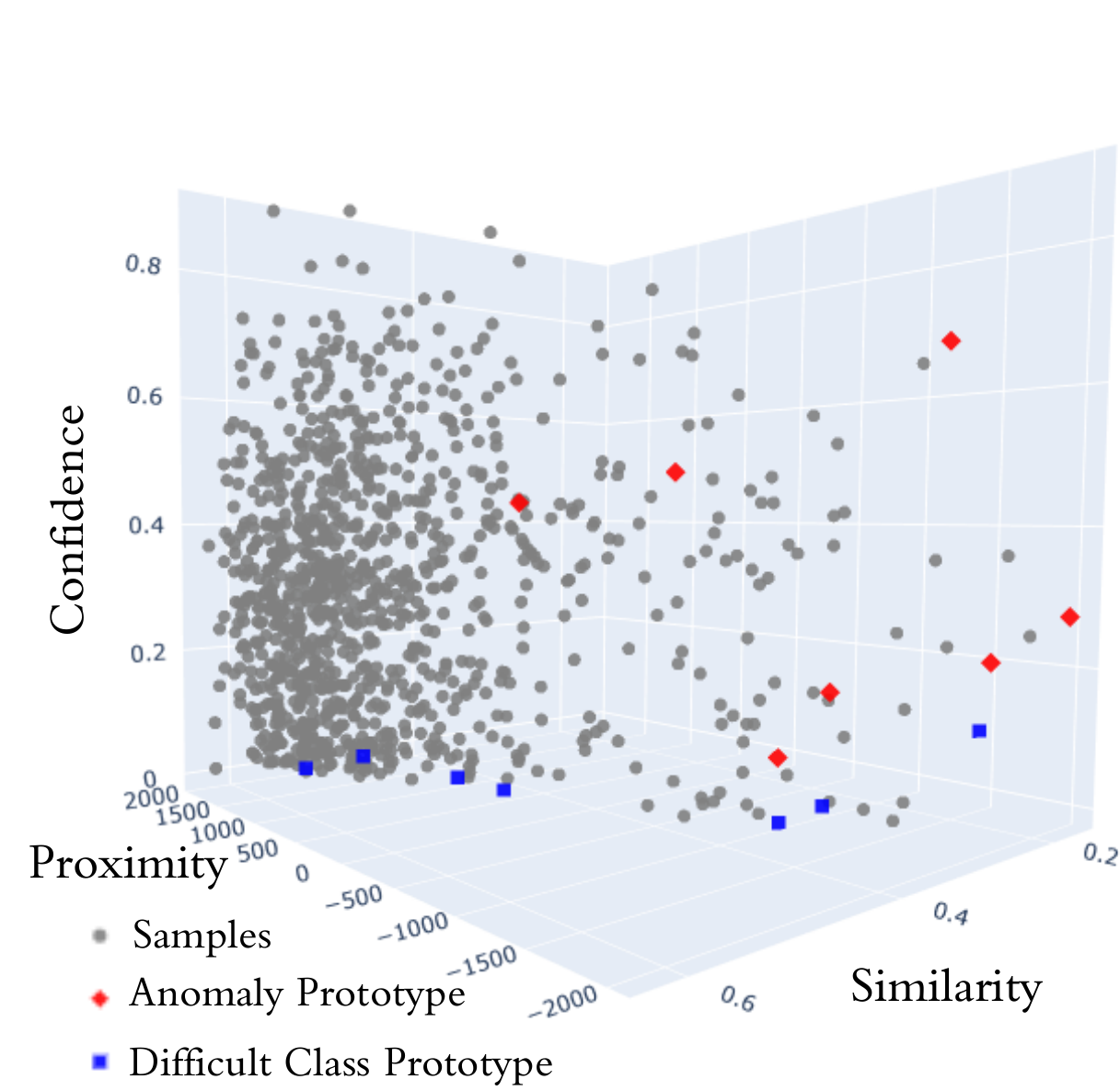}}}%
\qquad
 {{\includegraphics[trim={0.2cm 0.05cm 1.1cm 1cm},clip, width=.4425\linewidth]{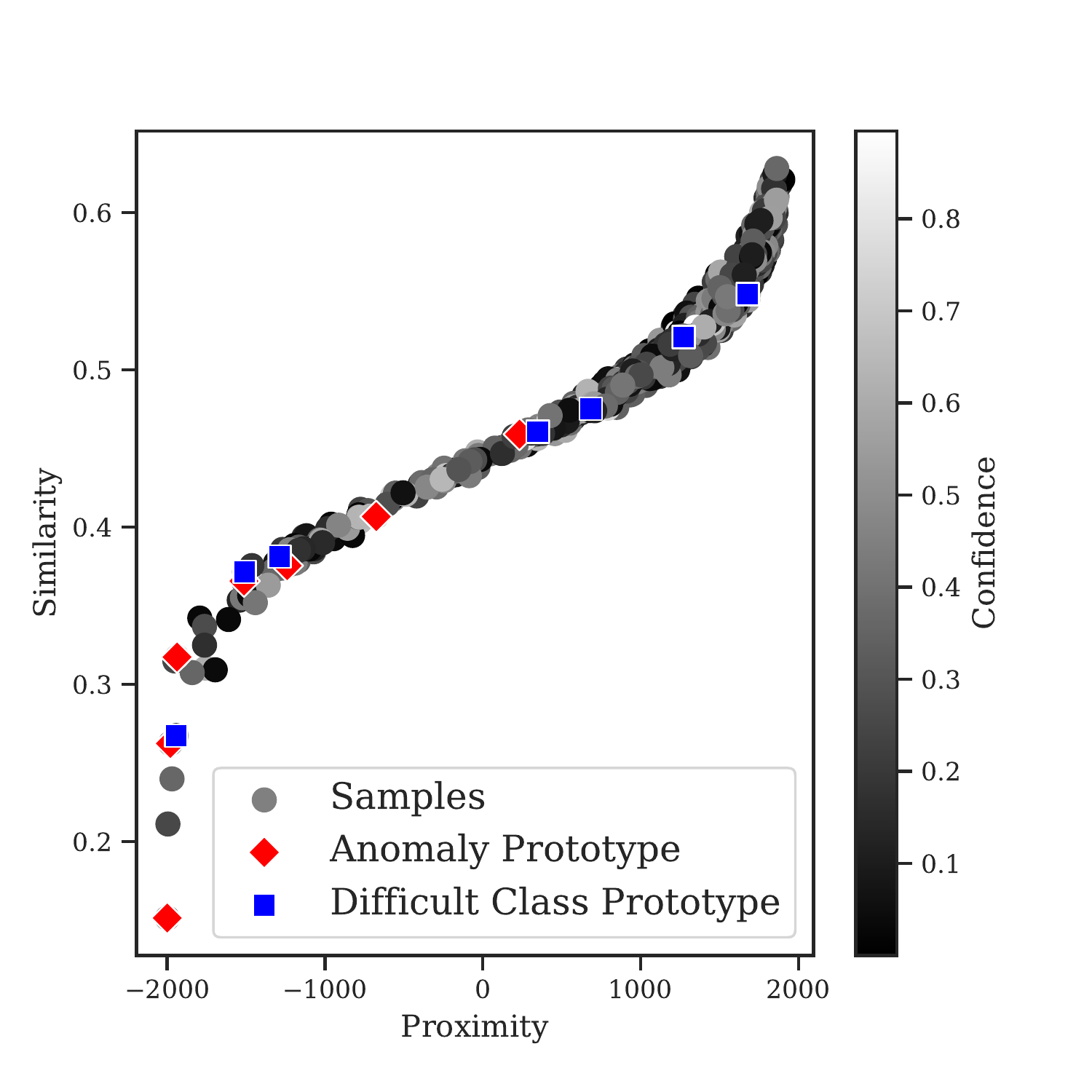} }}%
 \caption{Selected embedding prototypes of a single class of Twitter dataset. Difficult class prototypes have higher proximity, while anomaly prototypes suffer from low proximity due to their complex nature.}%
 \label{fig:protype_selection}
\end{figure}

We can now represent the embeddings in a three-dimensional space as shown in Figure \ref{fig:protype_selection} (similarity-proximity-confidence). 
The difficult class prototype selection follows three general rules: (i) it should prioritize low confidence samples (i) it should be `far' enough from existing prototypes (if any), (iii) it should have high `proximity' when possible. 
To this end, the first prototype with the lowest confidence, highest proximity, and highest similarity is chosen. Then, the subsequent difficult class prototypes are chosen in a logsparse~\cite{li2019enhancing} manner for every round with an exponential selection step of sample size ($\text{log}_{2}(N)$). 
Note that the samples are selected based on the low confidence, then high proximity but should have the lowest average similarity with the previously selected prototypes to be distinctive from each other. This strategy ensures us that the difficult class prototype are well represent problematic samples of the dataset.

Next, at a certain round ($t$), a prototype set $\mathbf{X}_{c}=\left\{\mathbf{x}_{c}^{(1)}, \ldots, \mathbf{x}_{c}^{(t)}\right\}$ is already formed for the $c$-th class, $c = 1,...,\Bar{c}$. 
Given any text $\mathbf{x}_i$, we can calculate the average cosine similarity between sample $\mathbf{x}_i$ and the selected prototype embeddings as:
\begin{equation} \label{eq:similar}
 s^{c}_{i,(c)}=\frac{1}{t} \sum_{j=1}^{t} s_{i,c^{(j)}},
\end{equation}
where $s^{c}_{i,(c)}$ is the average similarity of difficult class embeddings in the $j_{th}$ iteration for the $c$-th class. This average similarity can then be used as a complementary supervsion:

\begin{equation}
 {z^{c}_{i}}=\operatorname{argmax}_{c} \{s^{c}_{i,(c)}|c=1,...,\Bar{c}\}.
\end{equation}

As shown in Figure~\ref{fig:protype_selection}, difficult prototypes are chosen with low confidence levels, where they have the least similarity among the previously selected prototypes. During this process, we ensure that the subsequent prototypes stay far enough from existing prototypes so that there are limited redundant representations of the similar samples.

\subsection{Anomaly Prototypes}
\label{subsec:ap}

Anomaly prototypes are the selected sample prototypes that represent the scattered and shattered minority samples of a dataset. Such samples are often harder to detect and tend to deviate from normal samples. 

Given that the remaining classification error can be caused by such anomalies, it's important to not only capture such anomalies robustly but also leverage them for the optimization objectives of Transformers.

So far, difficult class prototypes can cover the problematic samples as they are detected by having high proximity and similarity.
However, a certain portion of prototypes may be located `far' from the difficult class prototypes and often represent the minority members of a class, as indicated by the red dots in the in Figure \ref{fig:protype_selection}. Such prototypes represent the minority of samples as they have a lower density.

The prototype with the least proximity $p_{min}$ is selected in the first round.
This ensures us that the elected prototype is representative of the minority samples. We then select the subsequent ones in the same logsparse manner as before, ensuring that the prototypes have the least similarity. 
The similarity score is calculated in a similar manner to Equation \eqref{eq:similar} while including the anomaly prototypes in the summation.

Figure \ref{fig:accuracy_improve} also illustrates the process, where gray dots represent all sample embeddings, and red dots indicate the embeddings of selected anomaly prototypes.

\begin{figure}[!t]
 \centering
 \includegraphics[trim={6.65cm 8cm 6.1cm 6.5cm},clip,width =1\linewidth]{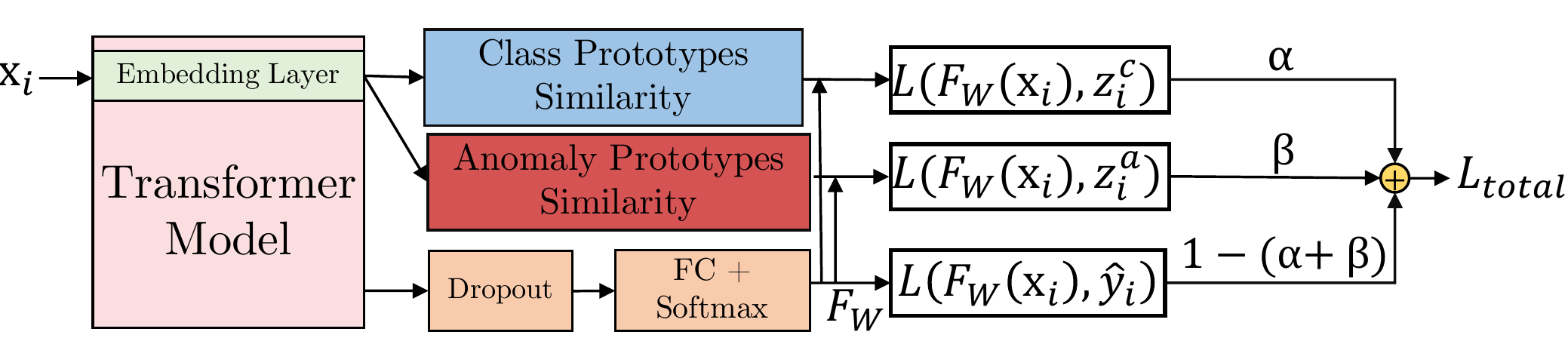}
 \caption{Protoformer leverages the embedding space to derive the difficult class and anomaly prototypes. The network is trained jointly on Transformer and similarity of embedding prototypes. The total loss is dependent
 on the $\mathbf{\alpha}$ and $\mathbf{\beta}$ values which are estimated in the training phase.}
 \label{fig:methodOverview}
  \vspace{-2mm}

\end{figure}

\subsection{Multi-Objective Self-Learning}
\label{subsec:self-learning}
Transformers used in text classification often rely on a single source of supervision which is the given labels. However, such design choice limits the Transformer's ability to perform well when the datasets are noisy labeled. Moreover, anomaly samples appear less in training compared to samples with high similarity. Note that majority of self-learning objectives for Transformers are to provide the greatest level of classification accuracy for all samples regardless of whether they are in the majority or minority. Intuitively, such a self-learning objective does not guarantee that the model suits well for minority classes due to their lower occurrence.
In order to incorporate our prototypes during the training and test stage, we introduce a multi-objective self-learning mechanism to Protoformer.

As shown in Figure~\ref{fig:methodOverview}, the similarities of embedding prototypes are used as self-supervision to train the Protoformer $\mathcal{F_W}$ after its fine-tuning state. The self-supervision is provided by the class prototype as below:
 \begin{equation}
 \mathcal{L}_{\text{proto}} = \frac{1}{m}\sum_{i=1}^m(\alpha \cdot \mathcal{L}(\mathcal{F}_{W}(\mathbf{x}_i),z^{c}_{i}) +\beta \cdot \mathcal{L}(\mathcal{F}_{W}(\mathbf{x}_i),z^{a}_{i})),
\end{equation}
where the weight factors $\alpha,\beta\in [0,1)$ and $\alpha+\beta < 1$ indicate the concentration of Transformer on the similarities of self-supervision of difficult class prototypes $z^{c}_{i}$ and anomaly prototypes $z^{a}_{i}$.
Hence, the overall loss is calculated by minimizing the classification loss based on three components:
\begin{equation} \label{eq:7}
 \mathcal{L}_{\text{total}} = (1-(\alpha+\beta))\cdot \frac{1}{m}\sum_{i=1}^m(\mathcal{L}(\mathcal{F}_{W}(\mathbf{x}_i),\hat{y}_{i}) + \mathcal{L}_{\text{proto}},
\end{equation}

To this end, when the network's predictions are in between classes, the network can improve its training by the self-supervision provided by the similarity of difficult class prototype $z^{c}_{i}$ and anomaly prototype $z^{a}_{i}$. Hence, we continue the training procedure iteratively until convergence:
$
W^{(t+1)} \leftarrow W^{(t)}-\xi \nabla\left(\mathcal{\mathcal{L}_{\text{total}}}\right),
$
where the gradient descent vector $\nabla(\mathcal{L}_{total})$ holds the partial derivatives of weights and biases of the total loss function, and $\xi$ is the learning rate.
We use a fully connected layer over
the final hidden state corresponding to the output token of the Transformer (i.e., \texttt{[CLS]} token). The softmax activation function is then applied to the hidden layer to provide classification.
It is important to note that this procedure can also be implemented solely during the test stage, which can make the calculation timing complexity of Protoformer similar to the fine-tuning process.

\subsection{Noisy Labels Enhancement}
\label{sec:noisy}
To mitigate the effect of noisy labels throughout the datasets, we are enhancing the labels through the similarities of embedding prototypes. This allows Protoformer to be robust toward datasets when the labels are not fully trustworthy. Consequently, when the labels are wrong, the training procedure of Transformers provides suboptimal weights, which makes the classification results untrustworthy.

Specifically, we can obtain the adjusted label of the a noisy labeled sample through maximum similarity to the difficult class prototype:
\begin{equation}
 {\Tilde{y}_i}=\operatorname{argmax}_{c} \{s_{i,(c)}|c=1,...,\Bar{c}\},
\end{equation}
where $s_{i,(c)}$ is the cosine similarity defined in Equation
\eqref{eq:similar} and the enhanced labels $\Tilde{y}$ can be used as a replacement for the noisy labels. Thus, the overall loss is calculated in a similar manner as Equation \eqref{eq:7}, while we are replacing the original noisy labels with the adjusted label.

\section{Experiments}
\label{section-experiment}
In this section, we provide descriptions for the datasets. We also describe the experimental settings and evaluation results. Lastly, we provide an analysis section that further discusses the effectiveness of Protoformer components.

\subsection{Benchmark Datasets \& Baselines} 

We have experimented with three challenging real-world datasets\footnote{Self-gathered datasets are accessible at 
https://github.com/ashfarhangi/Protoformer}.
The brief discussion for each dataset is as follows:

\begin{table}[!h]
 \centering
 \caption{Summary statistics of the evaluation dataset.}
\begin{tabular}{l|rrrrr} 
\hline
\textbf{Dataset} & \textbf{Twitter-Uni} & \textbf{IMDb} & \textbf{Arxiv-10} \\
\hline \textbf{\# Examples} & 25,000 & 25,000 & 100,000 \\
\textbf{\# Train} & 20,000 & 20,000 & 80,000 \\
\textbf{\# Validation} & 2,500 & 2,500 & 10,000 \\
\textbf{\# Test} & 2,500 & 2,500 & 5,000 \\

\textbf{\# Classes} & 8 & 2 & 10 \\

\hline
\end{tabular}
 
 \label{tab:dataset}
\end{table}

\textbf{Twitter-Uni$^{3}$.} We crawled over 12 million historical Tweets of 25,000 Twitter profiles from 8 U.S. college followers. 
As an example, the college of engineering holds near 3000 followers, which are labeled as engineering. Note that most existing benchmark Twitter datasets fail to hold high-quality labels that are provided by the original Twitter users. To alleviate this issue, we extracted a set
of students that stated their major in their Twitter bio. This set can serve as ground truth of the clean labels. 
We made this challenging new dataset available online, which can be used for future text classification or noisy label correction studies.

\textbf{ArXiv-10$^{3}$.} We also crawled the abstracts and titles of 100 thousand ArXiv scientific papers on ten research categories that include subcategories of computer science, physics, and math. The dataset is downsampled to contain exactly 10 thousand samples per category.

\textbf{IMDB.} The third dataset is the benchmark IMDb movie reviews~\cite{IMDBdataset}. The dataset is widely used as the sentiment classification task. It contains 25 thousand samples per sentiment (positive or negative). Both IMDb and ArXiv-10 datasets are originally labeled by the authors. It is however good to note that the labels are still susceptible to noisy labels.

The {\bf baseline} methods for comparison include:
\begin{itemize}
\item SVM~\cite{svm2003}, supervised learning with a linear separator to maximize the margin between classes, with the fine-tuned embeddings derived from the Transformers. 
\item HAN \cite{krishnan2017structured}, a hierarchical attention network for textual classification with word and sentence-level attention mechanisms. 
\item DocBERT \cite{adhikari2019docbert}, a document Transformer model with an LSTM architecture rather than a fully connected layer.
\item RoBERTa~\cite{liu2019roberta}, a Transformer with an improved
pretraining procedure. Specifically, showing improvement by removing the next sentence prediction pretraining objective.
\end{itemize}

\begin{table}[!h]
 \centering
 \caption{Hyperparameters of the Protoformer used for each dataset.} 
\begin{tabular}{lrrr}
\hline Parameter & Twitter-Uni & IMDb & Arxiv-10 \\
\hline Batch size & 32 & 64 & 32 \\
Learning rate & $5 \times 10^{-5}$ & $3 \times 10^{-5}$ & $5 \times 10^{-5}$ \\
Weight decay & $5 \times 10^{-5}$ & $1 \times 10^{-5}$ & $1 \times 10^{-4}$ \\
Preliminary training epochs & 5 & 3 & 2 \\
Fine-tuning epochs & 20 & 10 & 10 \\
Training time & 1:49h & 1:32h & 1:45h \\
Transformer & DistilBERT & BERT & RoBERTa \\

\hline
\end{tabular} 

 \label{tab:hyper}

\end{table}

\begin{table}[!t]
 \centering
\caption{Evaluation of the Protoformer and baseline methods. }

\begin{tabular}{l|rrr|rrr|rrr}

\hline
\hline
 & & Twitter & & & IMDb & & & ArXiv & \\
\hline
Model & Ma-F1 & Recall & Acc & Ma-F1 & Recall & Acc & Ma-F1 & Recall & Acc\\
 \hline SVM~\cite{svm2003} & 0.384 & 0.361 & 0.391 & 0.744 & 0.733 & 0.748 & 0.691 & 0.654 & 0.708 \\
HAN~\cite{yang2016hierarchical} & 0.412 & 0.392 & 0.425 & 0.894 &0.882 & 0.896 & 0.732 & 0.696 & 0.746 \\
DocBERT~\cite{adhikari2019docbert} & 0.521 & 0.506 & 0.534 & 0.932&0.921&0.936 &0.752&0.727&0.764 \\
RoBERTa~\cite{liu2019roberta} & 0.555 & 0.531& 0.567 & 0.952 &0.941 & 0.957 &0.769&0.732&0.779\\
Protoformer & \textbf{0.802} & \textbf{0.784}& \textbf{0.813} & \textbf{0.964} &\textbf{0.952}&\textbf{0.968} & \textbf{0.784} & \textbf{0.744} & \textbf{0.794}\\
\hline
\hline
\end{tabular}

\label{tab:results}
\end{table}

\subsection{Experimental Settings}
To showcase the generalization ability of our framework, we selected a unique Transformer for each dataset (Table~\ref{tab:hyper}).
The hyperparameters are based on the highest Macro-F1 score obtained on the validation set for all models (following the standard 80-10-10 split). We used a grid search approach to explore the hyperparameters: size of fully connected layer ${H}^{D} \in$ $\{256,512,768,1024\}$ and dropout $\delta \in\{0.0,0.1, \cdots, 0.9\}$. 
The experiments are conducted using PyTorch on a cloud workstation using Nvidia Tesla A100 GPU.

\subsection{Experimental Results}
For a less noisy labeled datasets such as IMDB and Arvix, the evaluated methods performed comparatively. Note that the majority of the classification error appears when the network does not show confidence in its classification, as was previously shown for the IMDB dataset in Figure~\ref{fig:analysis}. The Protoformer is also able to provide a competitive accuracy for cleaner datasets such as IMDb and ArXiv-10. Among the baselines, the performance of RoBERTa~\cite{liu2019roberta} is favorable compared to others. This is partly due to the different pretraining objectives from DocBERT.
As shown in Table~\ref{tab:results}, Protoformer resulted in the highest margin of accuracy for a noisy dataset, improving the Macro-F1 score from 55.5\% to 80.2\% for the Twitter-Uni dataset. We observed that this dataset provides the greatest difficulties for baseline methods where the models often misclassify problematic samples. To this end, we report a detailed accuracy breakdown for the Twitter dataset in Figure~\ref{fig:analysis}. The fine-tuning process for Transformers such as DocBERT, RoBERTa results in suboptimal classification.
Leveraging the selected prototypes, Protoformer was able to improve its classification accuracy on the harder and more complex samples (e.g., management students that are similar to other classes).
To this end, the fine-tuning process alone does not result in adequate accuracy due to the noise of the dataset. The combination of both embedding prototypes allows the Transformer to have a solution for anomalies and problematic samples of the dataset and further improves its generalization ability through difficult class prototypes.

\subsection{Analysis}
\label{section-analysis}
\begin{figure}[!t]%
 \centering
 {{\includegraphics[trim={1.5cm 1cm 2.08cm 1cm},clip,width =.31\linewidth] {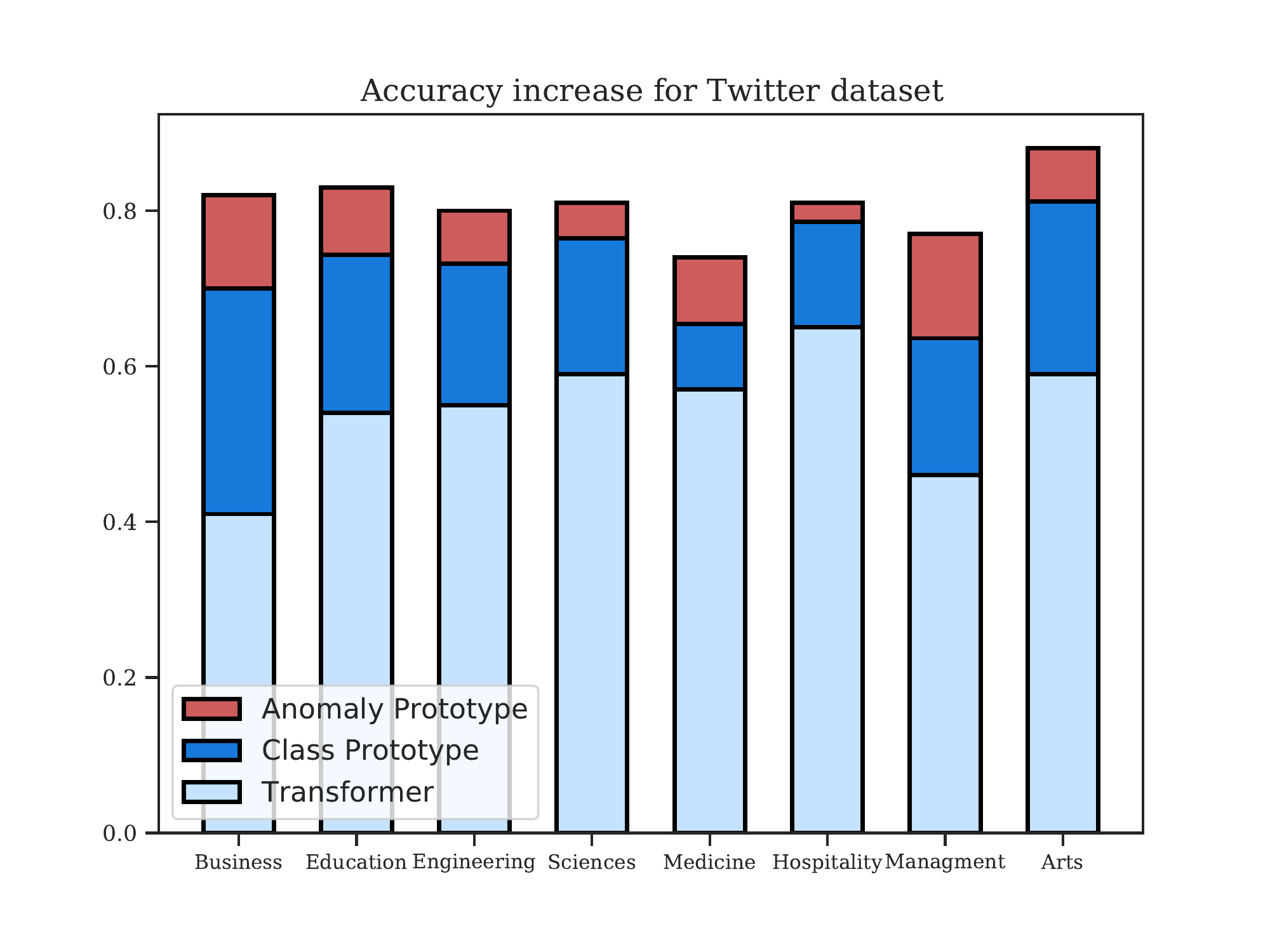}}}
 \qquad
 {{\includegraphics[trim={0.1cm 0cm 0.1cm 0cm},clip,width =.62\linewidth]{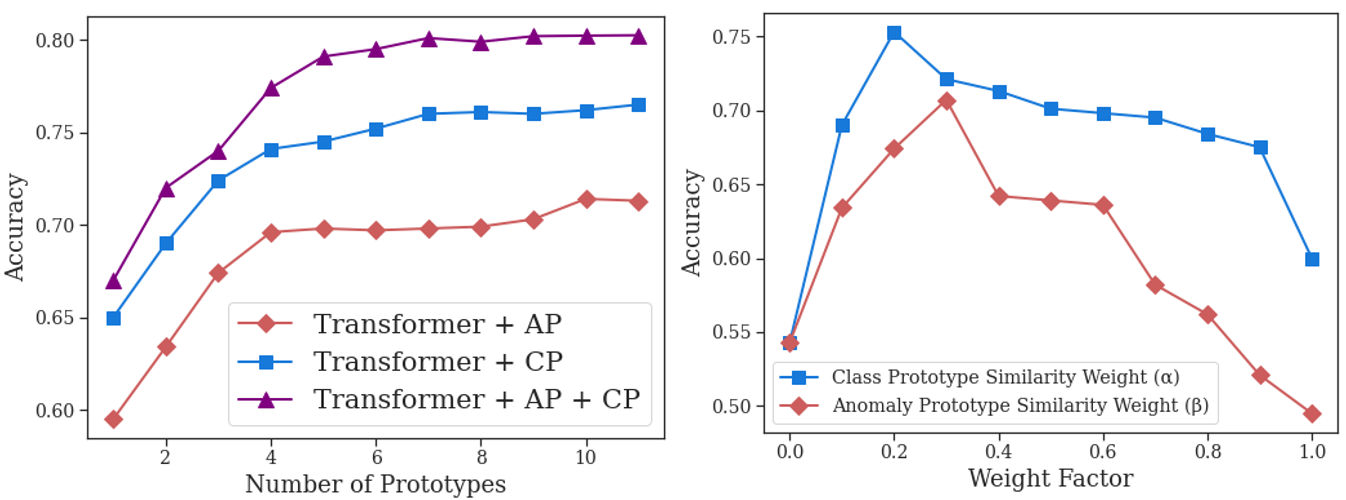}}}
 \caption{\textbf{Left:} Accuracy increase from the initial (light blue), class protoype (blue) and class anomalies (red), for Twitter dataset. Right: Influence of anomaly labeling of hurricanes for Collier county gross hotel sales revenue. \textbf{Middle:} Number of anomaly prototypes (AP) and difficult class prototypes (CP) per class for Twitter dataset. Higher number of prototypes resulted in marginal improvement while the combination of both category of prototypes gives us the optimal accuracy. \textbf{Right:} Testing accuracy with respect to the weight factors ($\alpha$ and $\beta$) ranging from 0 to 1.}%
 \label{fig:analysis}%
\vspace{-2mm}
\end{figure}

In this section, we provide an extensive analysis of the performance of Protoformer, as well as the role of each type of prototype on the overall performance. Hence, we limited the number of prototypes per class for the Twitter dataset and reported the changes. The results in Figure~\ref{fig:analysis} show that a single prototype is not sufficient to provide competitive accuracy even with the help of a fine-tuned Transformer. However, as the number of prototypes increased, we observed improvements in the accuracy of the Protoformer. The prototype selection procedure previously discussed ensures that there are multiple prototypes for every proximity metric, and the calculation of them is computationally expensive even for the large-scale dataset. Moreover, the weight factors are reported separately to showcase the effect of their self-supervision for the Twitter dataset. The results show that relying on the noisy labels ($\alpha$ and $\beta$ $= 0$) during training would be suboptimal and perform poorly on confirmed test data. 
Moreover, the accuracy would be optimal when weight factors sum to $0.5$ (i.e., $\alpha$=0.2, $\beta$$=0.3$).

\section{Conclusion}
\label{section-conclusion}
In this work, we developed a novel Transformer framework, Protoformer, that leverages the embedding prototypes of the dataset to enhance its generalization and specialization abilities.
It also includes a procedure for handling noisy labels. Various experiments are conducted to demonstrate the effectiveness of Protoformer over state-of-art topic and sentiment classification methods. 
For future work, we are interested in applying Protoformer for the image recognition tasks. We also like to explore the use of Protoformer on spherical and hyperbolic embedding space.

\vspace{-2mm}

\vspace{-2mm}
\subsection*{Acknowledgement}
Our work has been supported by the US National Science Foundation under grants No. 2028481, 1937833, and 1850851.

\bibliographystyle{splncs04}

\bibliography{Bibliography.bib}
\end{document}